\documentclass[runningheads,a4paper]{llncs}

\usepackage{amssymb}
\usepackage{graphicx}

\usepackage{hyperref}
\usepackage{url}
\usepackage{multirow}
\usepackage{graphicx}
\usepackage{subfig}
\usepackage{ntheorem}

\usepackage{algpseudocode,algorithm,algorithmicx}
\usepackage{amsmath,amssymb,amscd,amstext}
\usepackage{pgfplots}

\usepackage{url}
\urldef{\mailsa}\path|lgondara@sfu.ca|

\begin{document}
\mainmatter  

\title{Detecting Adversarial Samples Using Density Ratio Estimates}

\titlerunning{Detecting Adversarial Samples Using Density Ratio Estimates}

%
%
\author{Lovedeep Gondara}
\authorrunning{Lovedeep Gondara}

\institute{Simon Fraser University\\
\mailsa\\}

%
%

\toctitle{Lecture Notes in Computer Science}
\tocauthor{Authors' Instructions}
\maketitle

\begin{abstract}
Machine learning models, especially based on deep architectures are used in everyday applications ranging from self driving cars to medical diagnostics. It has been shown that such models are dangerously susceptible to adversarial samples, indistinguishable from real samples to human eye, adversarial samples lead to incorrect classifications with high confidence. Impact of adversarial samples is far-reaching and their efficient detection remains an open problem. We propose to use direct density ratio estimation as an efficient model agnostic measure to detect adversarial samples. Our proposed method works equally well with single and multi-channel samples, and with different adversarial sample generation methods. We also propose a method to use density ratio estimates for generating adversarial samples with an added constraint of preserving density ratio.
\end{abstract}

\section{Introduction}
\begin{figure}[t!] \centering
\begin{tabular}{cccc}
\subfloat[]{\includegraphics[scale=0.37]{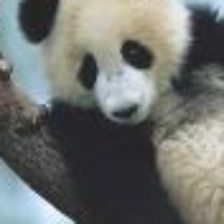}} &
\subfloat[]{\includegraphics[scale=0.37]{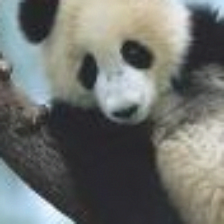}} &
\subfloat[]{\includegraphics[scale=0.18]{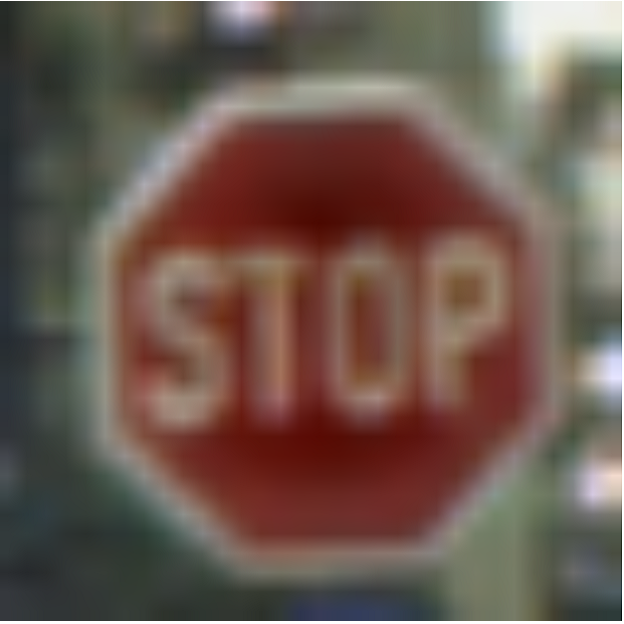}} & 
\subfloat[]{\includegraphics[scale=0.18]{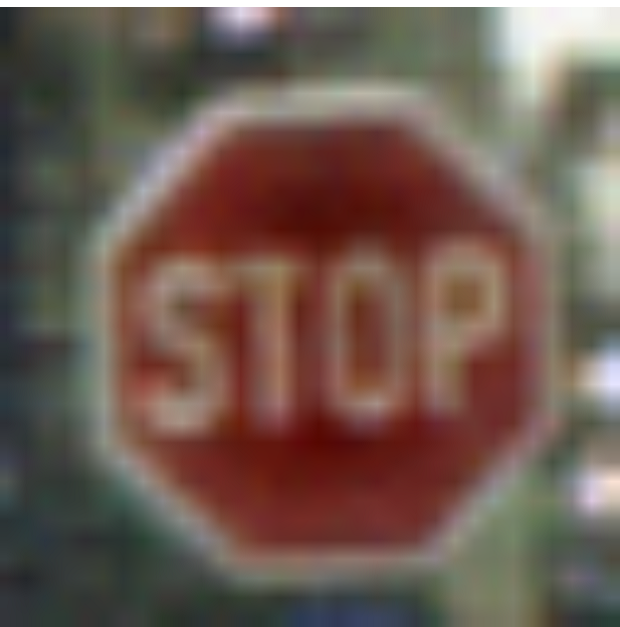}} 

\end{tabular}
\caption{Examples of adversarially generated images, (a) and (b) are from Goodfellow et al. \protect \cite{goodfellow2014explaining} where (a) is the original image classified correctly as "panda" by the model and (b) is  adversarially perturbed image, classified as "gibbon". Images (c) and (d) are from Papernot et al. \protect \cite{papernot2016practical} where (c) is the image of a stop sign correctly classified as a "stop sign" and (d) is adversarially perturbed image, classified as "yield" by the same model. }\label{figure1}
\end{figure}

Self driving cars \cite{bojarski2016end}, robotics \cite{levine2016learning}, computer games \cite{mnih2013playing}, imaging \cite{dahl2017pixel} and speech \cite{suwajanakorn2017synthesizing} are just some of the domains where deep learning has established state-of-the-art results and is being used in day to day applications. Recently \cite{goodfellow2014explaining} it has been established that such models can be easily fooled to misclassify with high probability using systematically perturbed inputs known as adversarial samples, which are visually imperceptible perturbations to real data. Existence of adversarial samples in real world poses a grave threat. Take an example of a self driving car, that uses a trained machine learning model to distinguish between different traffic signs. An adversarial sample can trick the model to classify stop sign as a yield sign \cite{papernot2016practical}, resulting in disastrous consequences. Figure \ref{figure1} shows two such examples, where models are tricked to incorrectly classify avdersarially generated images that are otherwise indistinguishable to human eyes from original images. Crafting effective defenses against adversarial samples is an important research problem and an active research area. However, effective solutions remain elusive due to the limited understanding of the nature of adversarial samples.

Adversarial samples, irrespective of the generative process, are perturbations to the original data. Assuming all data is generated from some underlying probability distribution admitting certain probability density, adversarial perturbations result in perturbed probability density regions. Our proposed method takes advantage the fact that we are able to detect such perturbations with high confidence. We propose to use direct density ratio estimation \cite{sugiyama2011least} as an intuitive, simple and model agnostic approach to effectively distinguish adversarial samples from real samples. To summarize, we make the following contributions:
\begin{enumerate}
    \item We present the first study proposing density ratio estimates as an efficient and model agnostic method for detecting adversarial samples.
    \item We show that adversarial detection based on density ratio estimates works for single and multi-channel samples alike (such as grayscale and colored images) and is transferable, that is, it works for different adversarial sample generation methods without the need of being trained for a specific one.
    \item We study the effective sample size required to estimate density ratio estimates so as to detect adversarial samples with high confidence in real life scenarios.
    \item We propose a modification to adversarial sample generation process by incorporating density ratio estimates, to generate adversarial samples that are closer to original samples with respect to their probability densities.
\end{enumerate}

Rest of the paper is organized as following: Next section presents preliminary introduction to direct density ratio estimation and methods for adversarial sample generation, followed  by the section \ref{sec:densityratio} presenting insights to how and why density ratio estimation works for detecting adversarial samples. In section \ref{sec:eval} we present empirical evaluation of our proposed method on MNIST and CIFAR-10 datasets followed by our proposed adversarial sample generation method in section \ref{sec:adv_gen}, related work in section \ref{sec:related} and finally we conclude the paper in section \ref{sec:conc}.

\section{Preliminaries} \label{sec:prelim}
This section gives preliminary introduction to direct density ratio estimation method and methods for generating adversarial samples.

\subsection{Density ratio estimation} \label{sec:density_ratio}
Comparing probability distributions is a primary task in statistical learning. One of the principled way of comparing two distributions $p_a(x), p_b(x)$, from datasets $a,b$, or the divergence between distributions is by estimating the density ratio:
\begin{equation}\label{ratio}
    r(x) = \dfrac{p_a(x)}{p_b(x)}
\end{equation}
A naive way of calculating $r(x)$ would be to explicitly estimate $p_a(x)$ and $p_b(x)$, and plug the estimates in \eqref{ratio}. But, direct density estimation is known to be a hard task \cite{vapnik1998statistical}. This is easily overcome by directly estimating $r(x)$, without having to estimate $p_a(x)$ and $p_b(x)$ separately \cite{kanamori2009efficient,nguyen2007estimating}. Different methods can be used to approximate $r(x)$, but in this paper we use the efficient unconstrained least squares importance fitting with cross-validation, details of which can be found in \cite{kanamori2009efficient}.

\subsection{Adversarial image generation}
Fast Gradient Sign Method (FGSM) for generating adversarial samples was introduced by Goodfellow et al. \cite{goodfellow2014explaining}. Given a model's cost function $c(M,x,y)$, adversarial sample is generated as $x^* = x + \sigma_x$, where $\sigma_x$ is computed as
\begin{equation}
    \sigma_x = \epsilon \, sign (\nabla_x c(M,x,y) )
\end{equation}
where $sign (\nabla_x c(M,x,y)$ is the sign of model's cost function gradient. $\epsilon$ controls the amount of perturbation, larger values of $\epsilon$ create highly perturbed images, distinguishable from real images by humans.

Jacobian Based Saliency Map Approach (JSMA) introduced by Papernot et al. \cite{papernot2016limitations} chooses perturbations by iteratively modifying only a limited number of features chosen based on decreasing adversarial saliency value where saliency values are calculated using model's Jacobian matrix.

Target Gradient Sign Methods (TGSM) works to create perturbations in a way so as to push the misclassifications towards a specific class \cite{kurakin2016adversarial}.

We refer readers to excellent works of Papernot et al. \cite{papernot2016practical} and Kurakin et al. \cite{kurakin2016adversarial} for further details.

\section{Density ratio estimation for detecting adversarial samples} \label{sec:densityratio}
This section explains why density ratio estimation is a good choice for detecting adversarial samples and how it is used in such scenarios.

\subsection{Intuition}
\begin{figure}\centering 
\begin{tabular}{cccc}
\subfloat[]{\includegraphics[scale=0.12]{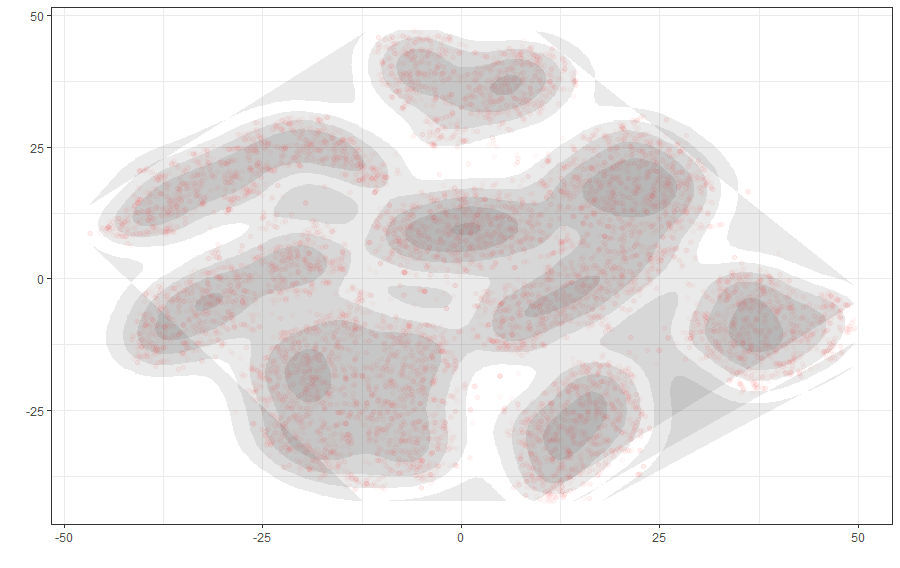}} &
\subfloat[]{\includegraphics[scale=0.12]{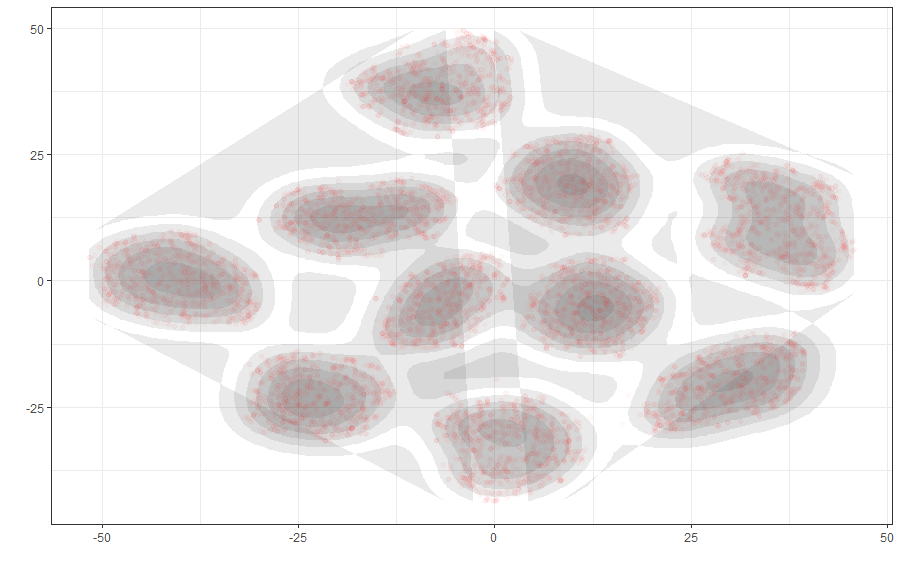}} &
\subfloat[]{\includegraphics[scale=0.12]{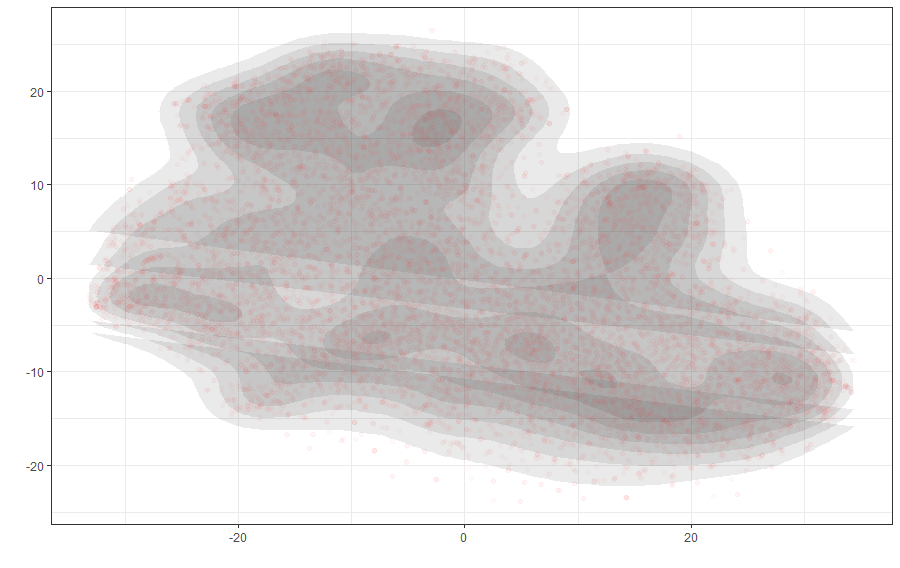}} & 
\subfloat[]{\includegraphics[scale=0.12]{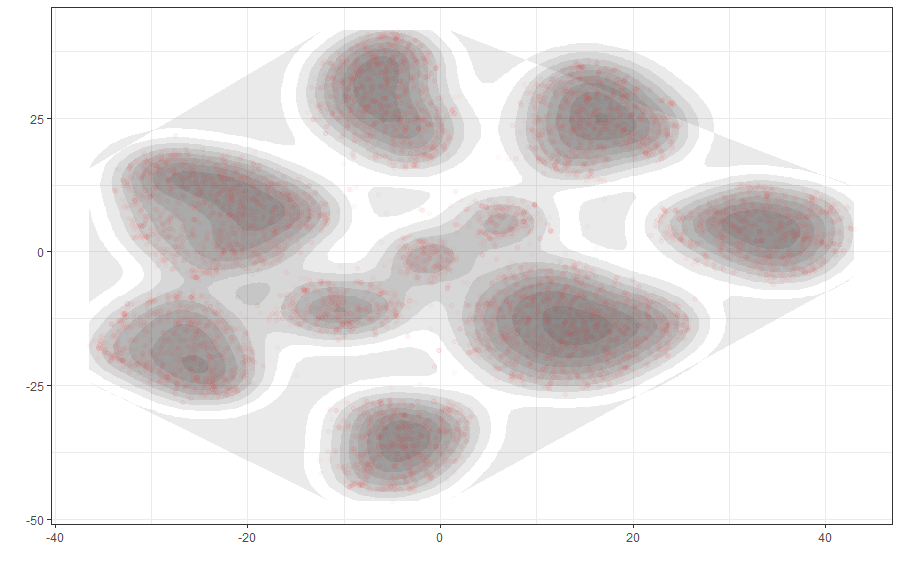}} 

\end{tabular}
\caption{tSNE plots for MNIST and grayscaled CIFAR-10 test datasets and their adversarial counterparts generated using FGSM with $\epsilon=0.3$. (a) is MNIST, real data; (b) is MNIST, adversarial data; (c) is CIFAR-10 grayscale, real data and (d) is it's adversarial version. It is clear that densities of real and adversarial data differ significantly.}\label{mnist_orig}
\end{figure}

All data are assumed to be generated from an underlying probability distribution with a certain probability density, approximable by a random sample. Creating adversarial samples involves perturbing original data in a systematic way. These perturbations lead to perturbed density regions, detected by direct density ratio estimation. Figure \ref{mnist_orig} explicitly shows this phenomenon, where we used tSNE \cite{maaten2008visualizing} to plot original and FGSM created adversarial MNIST \cite{lecunmnist2010} and grayscaled CIFAR-10 \cite{cifar10} images from their respective test partitions. We can clearly see the original images occupying certain density regions, which are very different from their adversarially generated counterparts. 

\subsection{Detection}
We start with a simple example, where we only have real images. That is, our dataset does not have any adversarial images. We begin by drawing two random samples of sufficiently large size from the dataset. For now, sufficiently large can be assumed to be $n=100$, and lets denote the two random samples by $X_1$ and $X_2$. Using unconstrained least squared approach, we estimate the density ratio of $X_1$ and $X_2$ as 
\begin{equation}\label{density_ratio}
    R(X)=\dfrac{p(X_1)}{p(X_2)}
\end{equation}
where $p(X_1)$ and $p(X_2)$ are probability densities of $X_1$ and $X_2$ respectively. If $X_1$ and $X_2$ are from the same underlying probability distribution, we can see that $R(X)$ would be approximated to be closer to 1. And, if the samples are from different distributions, the ratio would be farther away from 1. As should be the case with adversarial samples, which indeed is true as we show in later sections.

Readers might wonder: What if the images are colored? as this is the case in most real life scenarios. As a solution, we can simply extend our proposed method to estimate density ratio per color channel. In addition, we also define a density ratio estimate on the average of all three density ratio estimates of individual color channels, given as
\begin{equation}
     R(X_a) = \dfrac{1}{3} \Big (\dfrac{p(X_1^r)}{p(X_2^r)} + \dfrac{p(X_1^g)}{p(X_2^g)} + \dfrac{p(X_1^b)}{p(X_2^b)} \Big ) 
\end{equation}
where $X^r, X^g, X^b$ are red, green and blue channels respectively. The combined estimate can be used as a single statistic per comparison instead of using per channel individual estimates if required.

\section{Evaluation} \label{sec:eval}
This section presents empirical evaluation of our proposed method on colored and grayscale images with varying sample sizes and with varying adversarial sample generating methods.

\subsection{Experimental setup} \label{sec:exp_setup}
FGSM adversarial samples are generated using Cleverhans \cite{papernot2016cleverhans}, JSMA and TGSM samples are generated using Keras \cite{chollet2017} and TensorFlow \footnote{Code adapted from: \url {https://github.com/gongzhitaao/tensorflow-adversarial} }. Datasets MNIST \cite{lecun1998gradient} and CIFAR-10 \cite{krizhevsky2009learning} are the primary datasets used in this study, CIFAR-10 is used twice, once in its original form and second time in its grayscaled version. Test partitions of both datasets are used to generate adversarial samples. We define our setup for comparison using Algorithm \ref{density_comp}.
\begin{algorithm}  
  \caption{Density ratio based adversarial sample detection} \label{density_comp}
  \begin{algorithmic}   
    \Require $X$ and $Y$ as datasets with real and adversarial samples with $n$ samples in each.
      \For{$i \gets 1 \textrm{ to } t$}  
        \State $a$: Sample a random index of length $m$ without replacement from $n$
        \State $b$: Sample a random index of length $m$ without replacement from $n$
        \State $x$: Sample from real data using index $a$ = $X[a,]$
        \State $y$: Sample from adversarial data using index $b$ = $Y[b,]$
        \State $z$: Sample from real data using index $b$ = $X[b,]$
        \State Estimate density ratio $R_1$ = $\dfrac{p(x)}{p(y)}$
        \State Estimate density ratio $R_2$ = $\dfrac{p(x)}{p(z)}$
      \EndFor   
  \end{algorithmic}  
\end{algorithm}\\
To detect adversarial samples using density ratio estimation, we expect $R_1$ to be very different from $R_2$, where $R_2$ would be closer to 1. For initial comparisons, we keep $m$ and $t$ fixed at 100. That is, the experiments are run 100 times with sample size of 100 in the numerator and the denominator. Results are reported using mean $R_1$ and $R_2$ with related 95\% confidence intervals for statistical precision and comparison.

\subsection{Primary results}
We start with the standard evaluation on MNIST, grayscale version of CIFAR-10 and colored CIFAR-10 with adversarial samples generated using FGSM with varying values of $\epsilon$. 
\begin{table}[h!]
\centering
\caption{Density ratio estimates of real-adversarial and real-real samples. Real-real estimates are much closer to 1, as they should be compared to real-adversarial estimates. Average real-adversarial density estimates for all values of $\epsilon$ are significantly different from real-real estimates using mean comparison with $\alpha \le 0.05$. }
\label{grayscale-comp}
\begin{tabular}{l|l|l|l}
                            & $\epsilon$ & $\dfrac{Real}{Adversarial}$          & $\dfrac{Real}{Real}$        \\
                            &&&\\
                            \hline
\multirow{4}{*}{MNIST}      & 0.1        & 1.83(1.75,1.91)     & 1.32(1.23,1.41)  \\
                            & 0.3        & 20.70(20.43,20.96)  & 1.30 (1.21,1.39) \\
                            & 0.5        & 33.43(32.92,33.95)  & 1.32(1.23,1.41)  \\
                            & 1          & 34.15(33.60,34.70)  & 1.30(1.21,1.39)  \\
                                     \hline
\multirow{4}{*}{CIFAR-10(Grayscale)} & 0.1        & 1.37 (1.27,1.47)    & 1.02 (0.99,1.04) \\
                            & 0.3        & 30.21 (29.43,30.99) & 0.99(0.99,1.0)   \\
                            & 0.5        & 52.03 (50.28,53.78) & 1.00 (0.99,1.0)  \\
                            & 1          & 51.28(49.8,52.7)    & 1.01(0.99,1.03) \\
                            \hline
\multirow{4}{*}{CIFAR-10(Red)}        & 0.1        & 2.15(2.09,2.21)    & 0.99(0.99,1.0)  \\
                                  & 0.3        & 30.28(29.58,30.99) & 1.00(0.98,1.02) \\
                                  & 0.5        & 41.73(40.49,42.97) & 1.02(0.99,1.05) \\
                                  & 1          & 41.76(40.43,43.0)  & 1.00(0.99,1.00) \\
                                  \hline
\multirow{4}{*}{CIFAR-10(Green)}      & 0.1        & 1.90(1.80,2.0)     & 1.00(0.99,1.00) \\
                                  & 0.3        & 28.52(27.85,29.20) & 1.00(0.99,1.00) \\
                                  & 0.5        & 45.65(44.27,47.04) & 1.00(0.99,1.01) \\
                                  & 1          & 45.68(44.35,47.00) & 1.00(0.99,1.01) \\
                                  \hline
\multirow{4}{*}{CIFAR-10(Blue)}     & 0.1        & 1.66(1.54,1.77)    & 1.00(0.98,1.03) \\
                                  & 0.3        & 19.07(17.90,20.25) & 1.01(0.98,1.05) \\
                                  & 0.5        & 38.0(36.86,39.13)  & 1.04(1.0,1.09)  \\
                                  & 1          & 36.59(35.48,37.70) & 1.03(0.99,1.06) \\
                                 \hline
\multirow{4}{*}{CIFAR-10(Combined)} & 0.1        & 1.90(1.84,1.96)    & 1.00(0.99,1.01) \\
                                  & 0.3        & 25.96(25.21,26.71) & 1.00(0.99,1.02) \\
                                  & 0.5        & 41.79(40.99,42.59) & 1.02(1.00,1.04) \\
                                  & 1          & 41.34(40.52,42.16) & 1.00(0.99,1.02) \\
                                  \hline
\end{tabular}
\end{table}
Table \ref{grayscale-comp} shows the results, it is clear that real-real density ratios are closer to 1, as we would expect them to be, compared to real-adversarial estimates. It is also seen that as value of $\epsilon$ increases, the density ratio for real-adversarial samples deteriorates dramatically. Statistical tests for comparing means can be used to easily generate hypothesis and produce p-values for comparisons. In this scenario, all comparisons are statistically significantly different.

\subsection{Varying sample size}
Previously, we compared the density ratio estimates using a fixed sample size of 100 for real and adversarial images. In real life scenarios, we usually do not have 100 adversarial samples to estimate density ratio. So, to answer the question of how many samples are needed to reliably estimate density ratio and to reject adversarial samples with confidence. We use MNIST and CIFAR-10 (colored) to run density ratio estimations, but with varying sample sizes from 80 to 10, both for real and adversarial samples. That is, we simultaneously decrease sample size in the numerator and the denominator when estimating density ratios. We also decrease the noise parameter, $\epsilon$, to 0.1. Thus, creating adversarial samples that are closer to real samples and harder to detect.
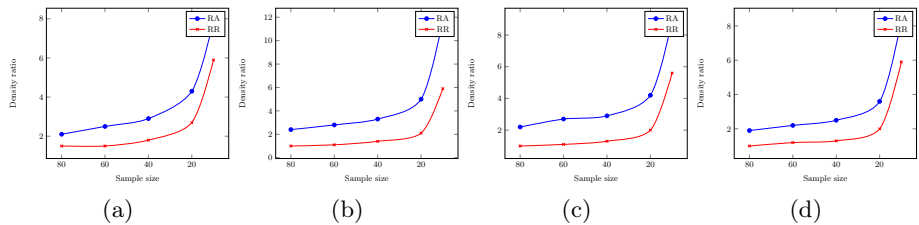
\begin{figure}[h]\centering
\begin{tabular}{cccc}
\subfloat[]
{\begin{tikzpicture}[scale=0.35]
    \begin{axis}[
        xlabel=Sample size,
        ylabel=Density ratio,
        x dir=reverse,
    ]
    \addplot[
        smooth,
        mark=*,
        blue,
        error bars/.cd, y dir=both, y explicit,
    ] plot coordinates {
        (10,7.9)
        (20,4.3)
        (40,2.9) 
        (60,2.5)
        (80,2.1)
    };
    \addlegendentry{RA}

    \addplot[smooth,color=red,mark=x]
        plot coordinates {
            (10,5.9)
            (20,2.7)
            (40,1.8) 
            (60,1.5)
            (80,1.5)
        };
    \addlegendentry{RR}
    \end{axis}
\end{tikzpicture}}
&
\subfloat[]
{\begin{tikzpicture}[scale=0.35]
    \begin{axis}[
        xlabel=Sample size,
        ylabel=Density ratio,
        x dir=reverse,
    ]
    \addplot[
        smooth,
        mark=*,
        blue,
        error bars/.cd, y dir=both, y explicit,
    ] plot coordinates {
        (10,11.7)
        (20,5.0)
        (40,3.3) 
        (60,2.8)
        (80,2.4)
    };
    \addlegendentry{RA}

    \addplot[smooth,color=red,mark=x]
        plot coordinates {
        (10,5.9)
        (20,2.1)
        (40,1.4) 
        (60,1.1)
        (80,1.0)
        };
    \addlegendentry{RR}
    \end{axis}
\end{tikzpicture}}
&
\subfloat[]
{\begin{tikzpicture}[scale=0.35]
    \begin{axis}[
        xlabel=Sample size,
        ylabel=Density ratio,
        x dir=reverse,
    ]
    \addplot[
        smooth,
        mark=*,
        blue,
        error bars/.cd, y dir=both, y explicit,
    ] plot coordinates {
        (10,8.9)
        (20,4.2)
        (40,2.9) 
        (60,2.7)
        (80,2.2)
    };
    \addlegendentry{RA}

    \addplot[smooth,color=red,mark=x]
        plot coordinates {
        (10,5.6)
        (20,2.0)
        (40,1.3) 
        (60,1.1)
        (80,1.0)
        };
    \addlegendentry{RR}
    \end{axis}
\end{tikzpicture}} 
&

\subfloat[]
{\begin{tikzpicture}[scale=0.35]
    \begin{axis}[
        xlabel=Sample size,
        ylabel=Density ratio,
        x dir=reverse,
    ]
    \addplot[
        smooth,
        mark=*,
        blue,
        error bars/.cd, y dir=both, y explicit,
    ] plot coordinates {
        (10,8.3)
        (20,3.6)
        (40,2.5) 
        (60,2.2)
        (80,1.9)
    };
    \addlegendentry{RA}

    \addplot[smooth,color=red,mark=x]
        plot coordinates {
        (10,5.9)
        (20,2.0)
        (40,1.3) 
        (60,1.2)
        (80,1.0)
        };
    \addlegendentry{RR}
    \end{axis}
\end{tikzpicture}} 

\end{tabular}
\caption{Density ratio estimates by varying sample size with epsilon=0.1, real-real (RR) density ratio is represented by red line and real-adversarial (RA) density ratio estimates are represented by blue line. (a) is MNIST, (b) is CIFAR red channel, (c) is CIFAR green channel, and (d) is CIFAR blue channel. It is seen that the density ratio estimates deteriorate with decreasing sample size, but a statistically significant difference persists between real-real and real-adversarial estimates.}\label{fig:sample_size}
\end{figure}\\    
Figure \ref{fig:sample_size} shows the results, it is seen that density ratio estimates deteriorate as sample size decreases, there is a noticeable shift from the starting sample size of 80 going to 10. But the difference between real and adversarial density ratio estimates persist, irrespective of number of samples. Using as few as 10 samples in the numerator and the denominator, it is possible to distinguish adversarial samples from real ones with high confidence.

\subsection{Transferability}
A significant implicit advantage of using density ratio estimates for the detection of adversarial samples is the tranferability. As real-real estimates are stationary within a sample, we can use the proposed method to detect adversarial samples generated using various different methods. Here, we generate adversarial samples using JSMA and TGSM. For detection, we use a fixed sample size of 100 and the comparison strategy described in section \ref{sec:exp_setup}.
\begin{table}[h]
\centering
\caption{Density ratio estimates of real-adversarial and real-real samples using dataset MNIST and colored CIFAR-10 with adversarial samples generated using JSMA and TGSM methods. It is seen that density ratio estimates are capable of detecting adversarial samples irrespective of adversarial sample generation process. Average real-adversarial density estimates for all values are significantly different from real-real estimates using mean comparison with $\alpha \le 0.05$.}
\label{extra-comp}
\begin{tabular}{l|l|l|l}
                            & $\epsilon$ & $\dfrac{Real}{Adversarial}$           & $\dfrac{Real}{Real}$         \\
                            &&&\\
                            \hline
\multirow{1}{*} {MNIST-JSMA} &         & 2.34(2.28,2.40)     & 1.10(1.01,1.16)  \\
\multirow{1}{*} {CIFAR-10-Red-JSMA} &         & 2.83(2.63,3.04)     & 1.00(0.98,1.02)  \\
\multirow{1}{*} {CIFAR-10-Green-JSMA} &         & 2.27(2.23,2.30)     & 1.00(0.99,1.01)  \\
\multirow{1}{*} {CIFAR-10-Blue-JSMA} &         & 2.25(2.14,2.35)     & 1.03(0.99,1.07)  \\
\multirow{1}{*} {CIFAR-10-Combined-JSMA} &         & 2.45(2.37,2.53)     & 1.01(1.00,1.03)  \\
                        
                                     \hline
\multirow{1}{*}{MNIST-TGSM} & 0.1        & 12.84(11.74,13.93)    & 1.06(1.01,1.11) \\
\multirow{1}{*}{MNIST-TGSM} & 0.05        & 6.87(6.80,6.95) & 1.09(1.04,1.15)   \\
\multirow{1}{*} {CIFAR-10-Red-TGSM} & 0.05        & 9.07(8.74,9.71)     & 1.03(0.99,1.06)  \\
\multirow{1}{*} {CIFAR-10-Green-TGSM} &   0.05      & 11.62(10.80,12.43)     & 1.02(0.99,1.05)  \\
\multirow{1}{*} {CIFAR-10-Blue-TGSM} &  0.05       & 9.34(8.85,9.83)     & 1.03(0.99,1.07)  \\
\multirow{1}{*} {CIFAR-10-Combined-TGSM} &  0.05       & 10.00(9.65,10.37)     & 1.03(1.01,1.05)  \\
                            
\end{tabular}
\end{table}\\
Table \ref{extra-comp} shows the results. It is seen that irrespective of adversarial sample generation process, adversarial samples have perturbed densities, making density ratio estimates an ideal candidate for detection with high confidence.

\subsection{A real life scenario}
So far we have demonstrated the capability of density ratio estimates for detecting adversarial samples with decreasing sample size in numerator and the denominator. In real life scenarios however, we always have enough supply of real images with constraints on availability of adversarial samples. Hence, here we investigate the effectiveness of using density ratio estimates to detect adversarial samples by only varying the sample size of adversarial images. As we have already shown the utility of our proposed method with sample size as small as 10, here we reduce the sample size for adversarial images even further, keeping the sample size for real images fixed at 100. This can be implemented as a slight modification to Algorithm \ref{density_comp}, where we change the length of random index $b$ keeping $a$ fixed at 100. As the results are similar on MNIST and CIFAR-10 datasets with different adversarial sample generation methods, here we only concentrate on samples generated using FGSM on MNIST data.
\begin{table}[]
\centering
\caption{Density ratio estimates of real-adversarial and real-real samples using dataset MNIST and FGSM created adversarial samples with varying adversarial sample size, $m$, and keeping sample size for real images fixed at 100. It is seen that even a single adversarial example can be detected using density ratio estimation with FGSM perturbation of $\epsilon=0.3$.}
\label{comp_app1}
\begin{tabular}{l|l|l|l}
                            & ($m$) & $\dfrac{Real}{Adversarial}$           & $\dfrac{Real}{Real}$         \\
                            &&&\\
                            \hline
\multirow{10}{*} {MNIST-FGSM} &   9      & 19.24(18.24,20.25)     & 1.55(1.43,1.66)  \\
                             &    8     & 19.84(18.84,20.84)     & 1.77(1.57,1.97)  \\
                             &    7     & 20.40(19.29,21.51)     & 1.99(1.66,2.33)  \\
                             &    6     & 19.46(18.33,20.59)     & 1.79(1.60,1.98)  \\
                             &    5     & 18.90(17.68,20.11)     & 1.87(1.63,2.11)  \\
                             &    4     & 17.53(16.22,18.85)     & 1.80(1.54,2.05)  \\
                             &    3     & 17.59(16.14,19.03)     & 2.15(1.57,2.73)  \\
                             &    2     & 19.90(18.53,21.26)     & 2.42(1.93,2.90)  \\
                             &    1     & 22.16(20.92,23.43)     & 6.48(5.18,7.77)  \\

\end{tabular}
\end{table}\\
We vary the sample size, $m$, of adversarial images from 9 to 1. That is, in a given density ratio estimation, we have 100 real samples with density $p(x)$ and $m$ adversarial samples with density $p(y)$. We also use $m$ real samples with density $p(z)$ for comparison of averaged $R_1$ and $R_2$. Table \ref{comp_app1} shows the results with adversarially created samples with $\epsilon=0.3$. It is seen that density ratio estimates are capable of detecting adversarial samples even when there is only one adversarial sample under investigation.

\section{Crafting adversarial samples using density ratio estimates}\label{sec:adv_gen}
We have demonstrated the capability of density ratio estimates to detect adversarial samples. In this section we study how we can leverage the density ratio estimates for crafting adversarial samples that explicitly preserve density ratio, while implicitly preserving probability density of original data. We initiate the sample crafting using the FGSM method, where we choose the epsilon based on the difference of density ratio estimates of real-real and real-adversarial samples. We define a tolerance parameter $\tau$ which is the desired/tolerated difference between real-real and real-adversarial density ratio estimates. Smaller $\tau$ means more similar probability densities. Process of finding optimal epsilon that satisfies our tolerance parameter is described in Algorithm \ref{density_craft}.
\begin{algorithm} 
  \caption{Crafting adversarial samples using density ratio estimates} \label{density_craft}
    \begin{algorithmic}
    \Require{$X$ as a dataset with real samples with $n$ observations; initial $\epsilon=\epsilon_i$, tolerance $\tau=\tau'$, step size for $\epsilon$: $\epsilon_{\Delta}$}
      \While{$\tau>\tau'$}
        \State $\epsilon_n = \epsilon_i - \epsilon_{\Delta}$
        \State $\epsilon = \epsilon_n$
        \State Generate adversarial batch with FGSM($\epsilon$)
            \For{$i \gets 1 \textrm{ to } t$}  
                \State $a$: Sample a random index of length $m$ without replacement from $n$
                \State $b$: Sample a random index of length $m$ without replacement from $n$
                \State $x$: Sample from real data using index $a$ = $X[a,]$
                \State $y$: Sample from adversarial data using index $b$ = $Y[b,]$
                \State $z$: Sample from real data using index $b$ = $X[b,]$
                \State Estimate density ratio $R_1$ = $\dfrac{p(x)}{p(y)}$
                \State Estimate density ratio $R_2$ = $\dfrac{p(x)}{p(z)}$
            \EndFor 
            \State Mean($R_1$)=$R_{1a}$
           \State Mean($R_2$)=$R_{2a}$
           \State $\tau$ = $|R_{1a}-R_{2a}|$
      \EndWhile
    \end{algorithmic}
\end{algorithm}
\begin{figure}[t!]\centering
\includegraphics[scale=0.25]{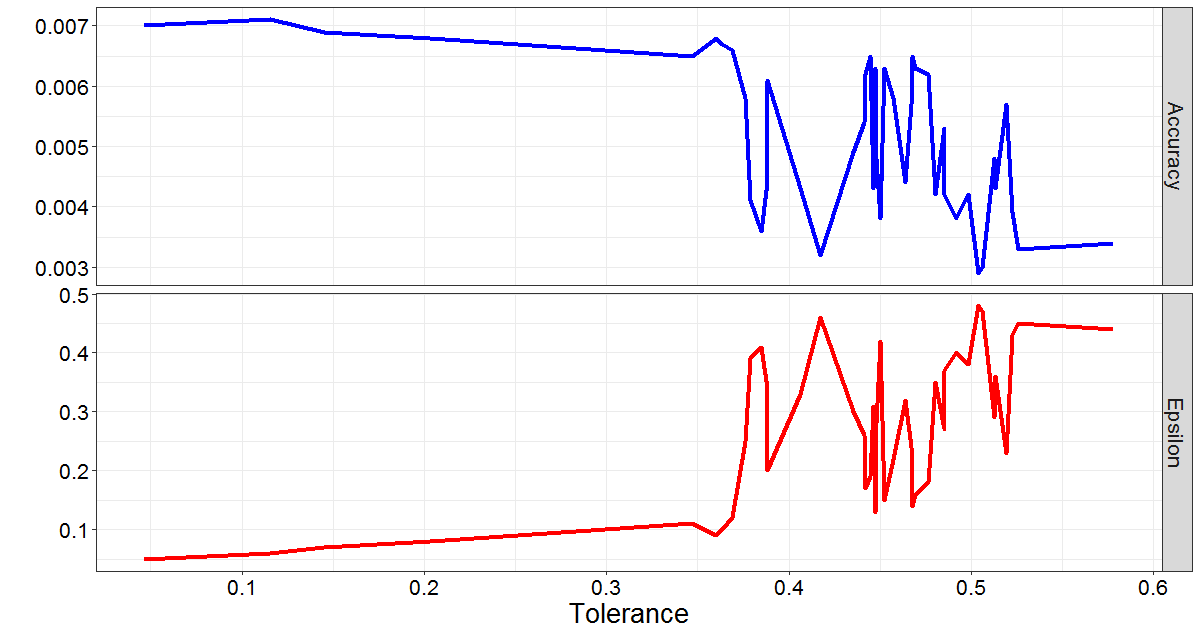}
\caption{Tolerance and epsilon,accuracy trade-off. Tolerance is on x-axis, smaller tolerance means more closer probability densities of real and adversarial samples. It is seen as the tolerance decreases, epsilon decreases and accuracy of the model increases, but not by a significant margin.}
\end{figure}\\
Figure \ref{density_craft} shows the trade-off of choosing tolerance parameter with respect to epsilon and classification accuracy. It is seen that as the tolerance is decreased, that is, adversarial and real samples are more closer with respect to density ratio estimates, value of epsilon decreases. It makes sense as lower epsilon means less perturbation. However, even decreasing the value of epsilon to be small enough so that the tolerance is 0.05 still yields effective adversarial samples, successful to fool the model without much increase in model's classification accuracy.

\section{Related work}\label{sec:related}
Crafting effective defences against adversarial attacks is an active area of research. Several defences against adversarial attacks have been proposed, such as defensive distillation \cite{papernot2016distillation} and training models using adversarial examples \cite{goodfellow2014explaining}. But they are generally computationally intensive and model specific, not agnostic  \cite{papernot2016practical}, similar to some other methods based on game theory \cite{liu2010mining,bruckner2011stackelberg}. Very recently, some interesting work has been done on adversarial sample detection, with Feinman et al. \cite{feinman2017detecting} working on detection of adversarial samples from artifacts, Li et al. \cite{li2016adversarial} using outputs from convolutional layers to detect adversarial samples and Grosse et al. \cite{grosse2017statistical} investigating the statistical detection of adversarial examples. Our work is closer to \cite{grosse2017statistical} where we both propose model agnostic adversarial sample detection methods. However, our methods are primarily different, ours is based on direct density ratio estimation compared to maximum mean discrepancy and we provide detailed insight to densities of adversarial samples with the utility of our method on multi-channel inputs.

\section{Conclusion}\label{sec:conc}
We have presented the evidence that adversarial samples have perturbed probability densities compared to real samples and it is possible to detect adversarial samples with high confidence using density ratio estimates. We have shown that density ratio estimates work well with single and multi-channel inputs, are capable of detecting adversarial samples generated using different methods, and with varying sample sizes. We have also presented a modification to the adversarial sample generation process, whereby incorporating density ratio estimate as a constraint, generating samples that are similar to original samples, not only in perception, but also with respect to probability densities.

\bibliographystyle{plain}
\bibliography{pakdd}

\begin{thebibliography}{10}

\bibitem{bojarski2016end}
Mariusz Bojarski, Davide Del~Testa, Daniel Dworakowski, Bernhard Firner, Beat
  Flepp, Prasoon Goyal, Lawrence~D Jackel, Mathew Monfort, Urs Muller, Jiakai
  Zhang, et~al.
\newblock End to end learning for self-driving cars.
\newblock {\em arXiv preprint arXiv:1604.07316}, 2016.

\bibitem{bruckner2011stackelberg}
Michael Br{\"u}ckner and Tobias Scheffer.
\newblock Stackelberg games for adversarial prediction problems.
\newblock In {\em Proceedings of the 17th ACM SIGKDD international conference
  on Knowledge discovery and data mining}, pages 547--555. ACM, 2011.

\bibitem{chollet2017}
François Chollet.
\newblock keras.
\newblock \url{https://github.com/fchollet/keras}, 2017.

\bibitem{dahl2017pixel}
Ryan Dahl, Mohammad Norouzi, and Jonathon Shlens.
\newblock Pixel recursive super resolution.
\newblock {\em arXiv preprint arXiv:1702.00783}, 2017.

\bibitem{feinman2017detecting}
Reuben Feinman, Ryan~R Curtin, Saurabh Shintre, and Andrew~B Gardner.
\newblock Detecting adversarial samples from artifacts.
\newblock {\em arXiv preprint arXiv:1703.00410}, 2017.

\bibitem{goodfellow2014explaining}
Ian~J Goodfellow, Jonathon Shlens, and Christian Szegedy.
\newblock Explaining and harnessing adversarial examples.
\newblock {\em arXiv preprint arXiv:1412.6572}, 2014.

\bibitem{grosse2017statistical}
Kathrin Grosse, Praveen Manoharan, Nicolas Papernot, Michael Backes, and
  Patrick McDaniel.
\newblock On the (statistical) detection of adversarial examples.
\newblock {\em arXiv preprint arXiv:1702.06280}, 2017.

\bibitem{kanamori2009efficient}
Takafumi Kanamori, Shohei Hido, and Masashi Sugiyama.
\newblock Efficient direct density ratio estimation for non-stationarity
  adaptation and outlier detection.
\newblock In {\em Advances in neural information processing systems}, pages
  809--816, 2009.

\bibitem{krizhevsky2009learning}
Alex Krizhevsky.
\newblock Learning multiple layers of features from tiny images.
\newblock 2009.

\bibitem{cifar10}
Alex Krizhevsky, Vinod Nair, and Geoffrey Hinton.
\newblock Cifar-10 (canadian institute for advanced research).
\newblock 2009.

\bibitem{kurakin2016adversarial}
Alexey Kurakin, Ian Goodfellow, and Samy Bengio.
\newblock Adversarial examples in the physical world.
\newblock {\em arXiv preprint arXiv:1607.02533}, 2016.

\bibitem{lecun1998gradient}
Yann LeCun, L{\'e}on Bottou, Yoshua Bengio, and Patrick Haffner.
\newblock Gradient-based learning applied to document recognition.
\newblock {\em Proceedings of the IEEE}, 86(11):2278--2324, 1998.

\bibitem{lecunmnist2010}
Yann LeCun and Corinna Cortes.
\newblock Mnist handwritten digit database.
\newblock 2010.

\bibitem{levine2016learning}
Sergey Levine, Peter Pastor, Alex Krizhevsky, Julian Ibarz, and Deirdre
  Quillen.
\newblock Learning hand-eye coordination for robotic grasping with deep
  learning and large-scale data collection.
\newblock {\em The International Journal of Robotics Research}, page
  0278364917710318, 2016.

\bibitem{li2016adversarial}
Xin Li and Fuxin Li.
\newblock Adversarial examples detection in deep networks with convolutional
  filter statistics.
\newblock {\em arXiv preprint arXiv:1612.07767}, 2016.

\bibitem{liu2010mining}
Wei Liu and Sanjay Chawla.
\newblock Mining adversarial patterns via regularized loss minimization.
\newblock {\em Machine learning}, 81(1):69--83, 2010.

\bibitem{maaten2008visualizing}
Laurens van~der Maaten and Geoffrey Hinton.
\newblock Visualizing data using t-sne.
\newblock {\em Journal of Machine Learning Research}, 9(Nov):2579--2605, 2008.

\bibitem{mnih2013playing}
Volodymyr Mnih, Koray Kavukcuoglu, David Silver, Alex Graves, Ioannis
  Antonoglou, Daan Wierstra, and Martin Riedmiller.
\newblock Playing atari with deep reinforcement learning.
\newblock {\em arXiv preprint arXiv:1312.5602}, 2013.

\bibitem{nguyen2007estimating}
XuanLong Nguyen, Martin~J Wainwright, and Michael~I Jordan.
\newblock Estimating divergence functionals and the likelihood ratio by
  penalized convex risk minimization.
\newblock 2007.

\bibitem{papernot2016cleverhans}
Nicolas Papernot, Ian Goodfellow, Ryan Sheatsley, Reuben Feinman, and Patrick
  McDaniel.
\newblock cleverhans v1.0.0: an adversarial machine learning library.
\newblock {\em arXiv preprint arXiv:1610.00768}, 2016.

\bibitem{papernot2016practical}
Nicolas Papernot, Patrick McDaniel, Ian Goodfellow, Somesh Jha, Z~Berkay Celik,
  and Ananthram Swami.
\newblock Practical black-box attacks against deep learning systems using
  adversarial examples.
\newblock {\em arXiv preprint arXiv:1602.02697}, 2016.

\bibitem{papernot2016limitations}
Nicolas Papernot, Patrick McDaniel, Somesh Jha, Matt Fredrikson, Z~Berkay
  Celik, and Ananthram Swami.
\newblock The limitations of deep learning in adversarial settings.
\newblock In {\em Security and Privacy (EuroS\&P), 2016 IEEE European Symposium
  on}, pages 372--387. IEEE, 2016.

\bibitem{papernot2016distillation}
Nicolas Papernot, Patrick McDaniel, Xi~Wu, Somesh Jha, and Ananthram Swami.
\newblock Distillation as a defense to adversarial perturbations against deep
  neural networks.
\newblock In {\em Security and Privacy (SP), 2016 IEEE Symposium on}, pages
  582--597. IEEE, 2016.

\bibitem{suwajanakorn2017synthesizing}
Supasorn Suwajanakorn, Steven~M Seitz, and Ira Kemelmacher-Shlizerman.
\newblock Synthesizing obama: learning lip sync from audio.
\newblock {\em ACM Transactions on Graphics (TOG)}, 36(4):95, 2017.

\bibitem{vapnik1998statistical}
Vladimir~Naumovich Vapnik and Vlamimir Vapnik.
\newblock {\em Statistical learning theory}, volume~1.
\newblock Wiley New York, 1998.

\end{thebibliography}
\end{document}